\documentclass[10pt,twocolumn,letterpaper]{article}
\pdfoutput=1

\usepackage{cvpr}
\usepackage{times}
\usepackage{epsfig}
\usepackage{graphicx}
\usepackage{amsmath}
\usepackage{amssymb}
\usepackage{graphicx}
\usepackage{tabularx}
\usepackage{float}

\usepackage[breaklinks=true,bookmarks=false]{hyperref}

\cvprfinalcopy 

\def\cvprPaperID{****} 

\begin{document}

\title{Automatic Pith Detection in Tree Cross-Section Images Using Deep Learning}

\author{Tzu-I Liao\\
Oregon State University\\
Corvallis, Oregon\\
{\tt\small zack.liao@oregonstate.edu}
\and
Mahmoud Fakhry\\
Oregon State University\\
Corvallis, Oregon\\
{\tt\small fakhryk@oregonstate.edu}
\and
Jibin Yesudas Varghese\\
Oregon State University\\
Corvallis, Oregon\\
{\tt\small jibinye@oregonstate.edu}
}
\maketitle

\begin{abstract}
  Pith detection in tree cross-sections is essential for forestry and wood quality analysis but remains a manual, error-prone task. This study evaluates deep learning models—YOLOv9, U-Net, Swin Transformer, DeepLabV3, and Mask R-CNN—to automate the process efficiently. A dataset of 582 labeled images was dynamically augmented to improve generalization.

Swin Transformer achieved the highest accuracy (0.94), excelling in fine segmentation. YOLOv9 performed well for bounding box detection but struggled with boundary precision. U-Net was effective for structured patterns, while DeepLabV3 captured multi-scale features with slight boundary imprecision. Mask R-CNN initially underperformed due to overlapping detections, but applying Non-Maximum Suppression (NMS) improved its IoU from 0.45 to 0.80.

Generalizability was next tested using an oak dataset of 11 images from Oregon State University's Tree Ring Lab. Additionally, for exploratory analysis purposes, an additional dataset of 64 labeled tree cross-sections was used to train the worst-performing model to see if this would improve its performance generalizing to the unseen oak dataset.   

Key challenges included tensor mismatches and boundary inconsistencies, addressed through hyperparameter tuning and augmentation. Our results highlight deep learning’s potential for tree cross-section pith detection, with model choice depending on dataset characteristics and application needs.

\end{abstract}

\section{Introduction}
\subsection{Background}
Pith detection in tree cross-sections is essential for forestry, dendrochronology, and wood manufacturing, providing valuable insights into tree growth patterns, wood quality, structural integrity, and historical climate conditions. The pith, located at the geometric center of the tree, acts as a crucial reference point for analyzing annual growth rings, detecting anomalies such as decay or reaction wood, and estimating mechanical properties that are vital for structural lumber grading.

Accurate pith localization significantly impacts industrial wood processing, particularly in sawmills, where precise pith detection facilitates optimal log cutting strategies, reduces wood waste, and improves the structural integrity and economic value of wood products. Furthermore, in dendrochronology, precise pith localization is pivotal for reconstructing past environmental and climatic conditions, as tree ring analyses depend heavily on accurately identifying the initial growth rings starting from the pith.

Historically, pith detection has been performed manually by skilled analysts using calipers, magnification tools, and image-processing techniques such as thresholding and edge detection. However, manual methods suffer from subjectivity, inconsistencies between analysts, and inefficiencies when processing large datasets. With the increasing demand for precision in forestry and wood engineering, there is a need for automated solutions that can deliver consistent, scalable, and accurate pith detection across diverse tree species and conditions.

Recent advancements in deep learning have demonstrated remarkable success in image-based segmentation and object detection tasks. Deep learning models, particularly convolutional neural networks (CNNs) and transformer-based architectures, have the potential to surpass traditional image processing methods in accuracy and efficiency. By leveraging these advancements, automated deep learning models can detect pith locations with higher consistency, reduced manual effort, and improved scalability across large datasets.

\subsection{Motivation}
The complexity of tree cross-sections poses a significant challenge for pith detection. Variations in wood grain patterns, knots, decay, and irregular growth rings make it difficult for traditional methods to generalize across different samples. Additionally, certain tree species exhibit non-uniform growth, leading to off-center or distorted pith locations, further complicating automated detection.

A major motivation for this study is to identify an optimal deep learning approach capable of handling these variations while maintaining high accuracy and generalization. While deep learning has seen widespread success in fields like medical imaging and autonomous navigation, its application to pith detection remains underexplored. By conducting a comparative study of different deep learning models, we aim to determine which approach is best suited for this task.

Furthermore, existing research often focuses on either object detection (bounding boxes) or segmentation (pixel-wise classification), but not both. Pith detection requires a hybrid approach—localizing the pith precisely while also segmenting its surrounding region to account for uncertainties in its structure. This study explores a diverse set of deep learning models to balance these requirements effectively.

From an industrial perspective, automated pith detection can enhance the efficiency of sawmill operations, where precise pith localization is required for optimal log cutting, defect detection, and wood classification. In research applications, more accurate pith detection can lead to better climate analysis through Dendrochronology studies and improved modeling of tree growth behaviors.

\subsection{Contributions}

To address the challenges associated with automated pith detection, this study presents a comprehensive evaluation of state-of-the-art deep learning models, explores data augmentation techniques to enhance generalization, optimizes segmentation performance using Non-Maximum Suppression (NMS), and employs rigorous performance metrics to assess model effectiveness.

A key contribution of this work is the comparative analysis of five deep learning models—\textbf{Swin Transformer, YOLOv9, DeepLabV3, U-Net, and Mask R-CNN}—each offering distinct architectural advantages and trade-offs in terms of segmentation quality, detection speed, and computational efficiency. By systematically evaluating these models on a dataset of tree cross-sections, we provide insights into their suitability for pith detection, identifying the most effective approach while considering performance-cost trade-offs. This comparative study offers valuable guidance for selecting the optimal deep learning framework for automated pith localization.

To mitigate the challenge posed by limited labeled datasets, this study implements a dynamic data augmentation pipeline that enhances dataset diversity on-the-fly. The augmentation techniques employed include \textbf{random horizontal and vertical flips}, which account for natural variations in image orientation, and \textbf{rotations of up to $\pm30^\circ$}, simulating different cutting angles of tree cross-sections. Additionally, \textbf{color jittering}, involving adjustments to brightness, contrast, saturation, and hue, is introduced to improve robustness against lighting variations. \textbf{Scaling and cropping} further enhance the model’s ability to generalize across different tree sizes and structures. By increasing dataset variability without inflating storage requirements, these augmentation strategies enable the models to learn more generalized features, reducing the risk of overfitting and improving predictive reliability across diverse input conditions.

Another important contribution of this study is the optimization of \textbf{Mask R-CNN} through the implementation of \textbf{Non-Maximum Suppression (NMS)}. During initial evaluations, Mask R-CNN exhibited performance degradation due to overlapping bounding boxes around the pith region, leading to inconsistent and redundant predictions. To address this issue, NMS was introduced to filter redundant bounding boxes based on their Intersection over Union (IoU) scores. By applying an IoU threshold of \textbf{0.5}, the model’s IoU score improved significantly from \textbf{0.45 to 0.80}, resulting in more precise and reliable pith localization. This enhancement demonstrates the importance of post-processing techniques in improving segmentation accuracy, particularly in complex detection tasks where bounding box overlap is prevalent.

Finally, this study employs a rigorous evaluation framework using multiple performance metrics to comprehensively assess model effectiveness. These metrics include \textbf{accuracy}, which measures the overall correctness of pith localization, and \textbf{precision}, which evaluates the proportion of correctly detected pith locations. Additionally, \textbf{recall} is used to quantify the model’s ability to detect true pith locations, while the \textbf{F1-score} provides a balanced measure of precision and recall. The \textbf{Intersection over Union (IoU)} metric is employed to assess the overlap between predicted and ground truth pith locations, providing a more granular evaluation of segmentation performance. By analyzing these metrics across different models, this study offers a detailed performance breakdown, highlighting the trade-offs between detection accuracy, processing speed, and segmentation quality, thus facilitating informed decision-making for future applications in forestry and wood science.

\section{Literature Review}

\subsection{Traditional Methods}

Historically, pith detection has been performed manually using calipers and image-processing techniques such as thresholding, contour detection, and edge-based methods \cite{schraml2013pith, norell2009automatic}. While effective in controlled environments, these methods are highly sensitive to variations in wood structure, lighting conditions, and surface textures, leading to inconsistencies and a lack of robustness when applied to diverse datasets.

Schraml and Uhl \cite{schraml2013pith} introduced a Local Fourier Spectrum Analysis (LFSA) approach, which segments the wood cross-section into patches and estimates each patch’s orientation using a 2D Fourier Transform. A Hough Transform-based accumulation method was then used to localize the pith. Although this approach proved effective for well-defined tree rings, it struggled in cases with irregular growth patterns, making it less adaptable to natural variations in wood structures. Similarly, Kurdthongmee et al. \cite{kurdthongmee2018fast} utilized Histogram of Oriented Gradients (HOG) for tree ring orientation estimation. While this method relied on predefined rules for accumulating orientations, it demonstrated lower robustness when dealing with noise and missing ring structures.

Norell et al. \cite{norell2009automatic} proposed the use of quadrature filters and Laplacian pyramids to extract multi-scale edge features, aiming to enhance pith detection robustness. Despite improvements in feature extraction, this method still lacked the ability to generalize across different wood species and imaging conditions. More recently, Decelle et al. \cite{decelle2022ant} introduced an Ant Colony Optimization (ACO) algorithm to refine pith localization. Although the algorithm showed strong performance on structured datasets, its computational inefficiency limited its practicality for real-time applications.

\subsection{Deep Learning Approaches}

With the advancement of deep learning, researchers have begun to explore the application of Convolutional Neural Networks (CNNs) and transformer-based architectures for wood structure analysis. Compared to traditional approaches, deep learning models offer higher accuracy, better generalization, and increased automation, making them promising solutions for pith detection.

CNN-based methods have demonstrated significant success in pith detection by learning hierarchical feature representations. One of the widely adopted deep learning approaches is the You Only Look Once (YOLO) framework, a real-time object detection model optimized for bounding box-based tasks. Kurdthongmee et al. \cite{kurdthongmee2020comparative} conducted a comparative study of YOLOv3 and SSD MobileNet for pith detection in sawmill environments, showing that YOLO performed well in detecting pith locations efficiently. Another common architecture used for segmentation tasks is U-Net, originally designed for medical image segmentation. U-Net employs an encoder-decoder structure that enables the model to learn fine-grained segmentation features, making it particularly suitable for wood cross-sections \cite{ronneberger2015u}. Additionally, Mask R-CNN, an extension of Faster R-CNN, has been used for object detection and segmentation. Unlike conventional bounding-box-based models, Mask R-CNN generates segmentation masks alongside object detection, allowing for more precise localization of the pith. In this study, the Mask R-CNN segmentation mask was disabled, reducing the model's complexity to that of Faster R-CNN. Additionally, since studies have shown that overlapping bounding boxes in Mask R-CNN can negatively impact performance in pith detection, this motivated the use of Non-Maximum Suppression (NMS) to filter redundant predictions \cite{he2017mask}. 

More recently, transformer-based models have demonstrated improved segmentation accuracy by capturing long-range dependencies in images. The Swin Transformer has emerged as a powerful alternative to CNNs for segmentation tasks. Unlike traditional convolutional networks, Swin Transformers utilize hierarchical feature learning, making them particularly effective in segmenting pith regions at varying scales \cite{liu2021swin}. Another prominent model, DeepLabV3, is designed for semantic segmentation and incorporates Atrous Spatial Pyramid Pooling (ASPP) to capture multi-scale context \cite{chen2017deeplab}. This capability enhances the model’s ability to detect complex textures and irregular growth rings, making it a promising approach for pith localization.

\subsection{Comparison of Traditional vs. Deep Learning Approaches}

Traditional methods of pith detection rely heavily on handcrafted feature extraction techniques, which can be effective in well-controlled environments but often lack adaptability to variable conditions such as differences in lighting, wood species, and structural anomalies. These methods require extensive fine-tuning and are prone to inconsistencies when applied to diverse datasets. In contrast, deep learning models automatically learn feature representations, leading to improved generalization across different input conditions.

Despite their advantages, deep learning approaches also have certain limitations. One of the key challenges is their high computational requirements, particularly for transformer-based models, which demand substantial processing power and memory. Additionally, these models are highly sensitive to training data, requiring large, well-annotated datasets to achieve robust performance. Without sufficient data, deep learning models are prone to overfitting, particularly in scenarios where training samples are limited.

Building on prior research, this study presents a comparative analysis of five deep learning models—YOLOv9, U-Net, Swin Transformer, DeepLabV3, and Mask R-CNN—while also integrating dynamic data augmentation techniques to enhance model generalization. Furthermore, this study addresses specific limitations in segmentation performance by implementing Non-Maximum Suppression (NMS) and optimizing hyperparameters to mitigate the effects of overlapping detections and boundary inconsistencies. By systematically evaluating these approaches, this research aims to identify the most effective model for automated pith detection in tree cross-sections.
\section{Methodology}
\subsection{Dataset}
The dataset used in this study consists of 582 labeled tree cross-section images, each annotated with the ground truth pith location. These images were sourced from forestry research databases and publicly available datasets \cite{DBLP:journals/corr/abs-2404-01952}. All images were resized to 640 $\times$ 640 pixels during preprocessing. Given the limited size of the dataset, data augmentation techniques were applied to enhance model generalization and ensure robustness across varying tree species and structural differences. The augmentation pipeline included random horizontal and vertical flips to account for different tree orientations, rotations of $\pm30^\circ$ to simulate variations in cutting angles, color jittering to introduce variations in illumination conditions, and scaling and cropping to improve the model’s ability to generalize across different tree sizes.

\begin{table}[htbp]
\centering
\resizebox{\linewidth}{!}{
\begin{tabular}{|l|c|l|}
\hline
\textbf{Collection} & \textbf{Size} & \textbf{Tree Specie} \\ 
\hline
UruDendro2 & 119 & \textit{Pinus taeda} \\ 
UruDendro3 & 9 & \textit{Gleditsia triacanthos} \\ 
Kennel & 7 & \textit{Abies alba} \\ 
Forest & 57 & \textit{Douglas fir} \\ 
Logyard & 32 & \textit{Douglas fir} \\ 
Logs & 150 & \textit{Douglas fir} \\ 
Discs & 208 & \textit{Douglas fir} \\ 
\hline
Total & 582 & \\
\hline
\end{tabular}}
\caption{Dataset description}
\label{tab:dataset_description}
\end{table}

The dataset was split into training, validation, and testing subsets, with 70\% of the data allocated for training, 15\% for validation, and 15\% for testing. This distribution ensured that the models were trained on a diverse set of images while maintaining sufficient data for evaluating generalization performance.

The code for all trained models was stored in the following GitHub repository: \texttt{github.com/Mahmoud}
\texttt{Fakhry/Automatic\_Pith\_Detection}.

\begin{figure}[t]
    \begin{center}
        \fbox{\includegraphics[width=0.9\linewidth]{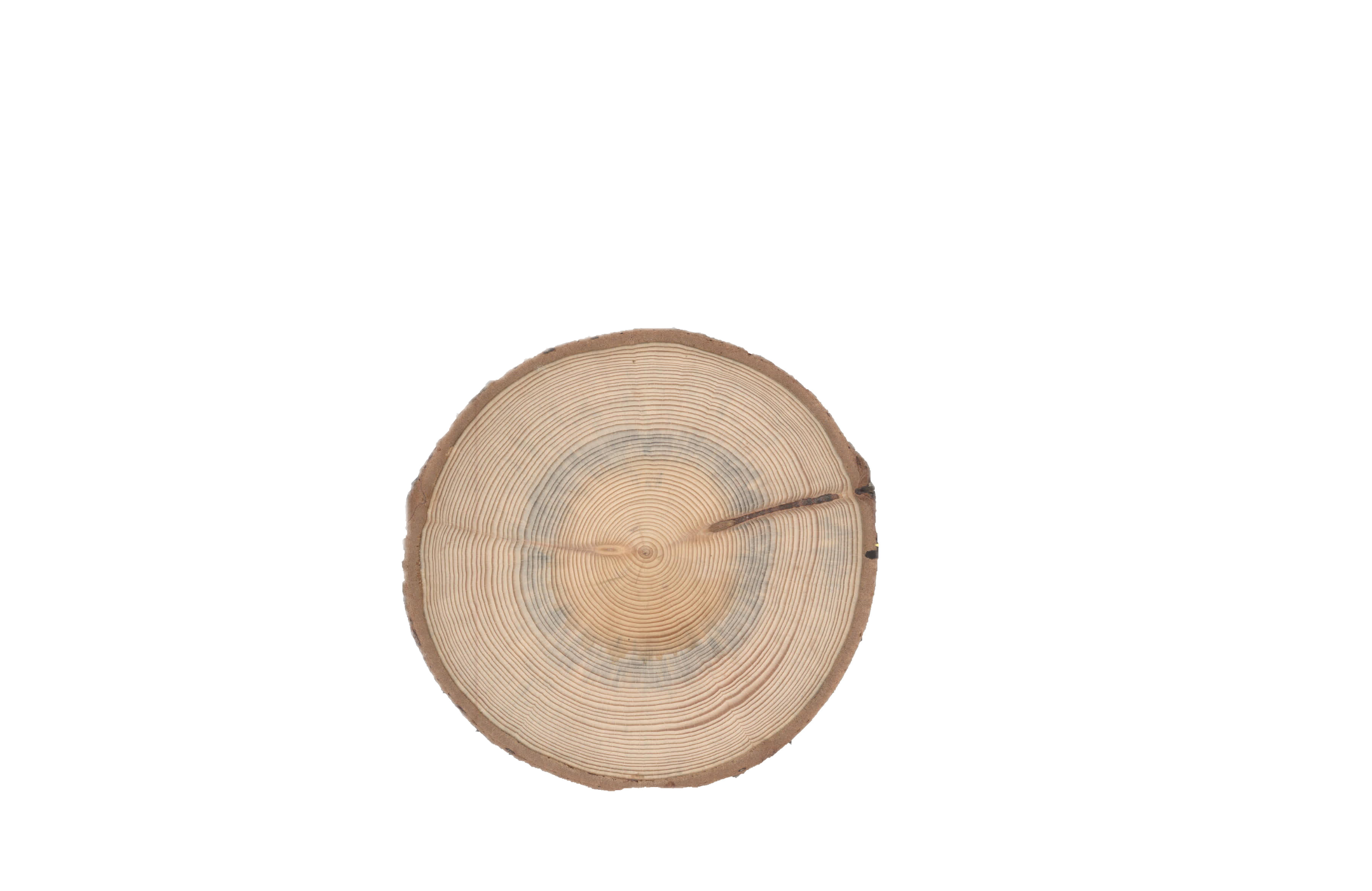}}
    \end{center}
    \caption{Example of an input tree cross-section image from the dataset.}
    \label{fig:input_example}
\end{figure}

\subsection{Model Architectures and Training}
This study investigates five different deep learning models—Swin Transformer, YOLOv9, DeepLabV3, U-Net, and Mask R-CNN—each selected based on their ability to perform object detection and segmentation tasks. The models were trained on the augmented dataset using specific hyperparameter configurations optimized to improve performance.

\subsubsection{Swin Transformer}
The Swin Transformer was selected for its ability to efficiently capture both local and global image features through hierarchical shifted window attention mechanisms. Unlike conventional CNNs, which rely on fixed receptive fields, the Swin Transformer processes images in a hierarchical manner, progressively increasing the window size, making it particularly effective in segmenting tree cross-sections with complex structures. The model was trained with a batch size of 8, a learning rate of 0.0001 using the Adam optimizer, and cross-entropy loss as the objective function. Despite its superior segmentation capabilities, Swin Transformer has a high computational cost, requiring significant memory for training and inference.

\subsubsection{YOLOv9}
YOLOv9 was included in this study due to its efficiency in real-time object detection. Its single-stage architecture enables fast inference, making it a viable choice for industrial applications requiring rapid pith detection. YOLOv9 was trained using a batch size of 8, a learning rate of 0.0001 with the Adam optimizer, and an object detection loss function based on IoU. Although YOLOv9 demonstrated excellent speed, it struggled with fine boundary details, particularly in cases where the pith was irregularly shaped or had diffuse edges.

\subsubsection{DeepLabV3}
DeepLabV3 was selected for its ability to perform multi-scale feature extraction using Atrous Spatial Pyramid Pooling (ASPP). This model was particularly effective in detecting pith locations across varying tree species, capturing both fine and coarse features simultaneously. The model was trained with a batch size of 6, a learning rate of 0.0001, and the StepLR learning rate scheduler with a step size of 3 and gamma value of 0.1. The Adam optimizer was used, and cross-entropy loss served as the objective function. While DeepLabV3 achieved robust segmentation results, it exhibited slight boundary imprecision in high-noise images.

\subsubsection{U-Net}
U-Net, originally designed for biomedical image segmentation, was included due to its strong encoder-decoder structure, which enables effective segmentation in structured images. The model was trained with a batch size of 8, a learning rate of 0.0001 using the Adam optimizer, and binary cross-entropy loss. U-Net was computationally lightweight and performed well in segmenting well-defined pith regions. However, it struggled with complex textures, leading to occasional misclassifications in tree cross-sections with non-uniform growth patterns.

\subsubsection{Mask R-CNN}
The Mask R-CNN used was the TorchVision Mask R-CNN, which was pretrained on the COCO dataset and uses a ResNet-50 backbone with a Feature Pyramid Network (FPN). The implementation used disabled the generation of segmentation masks, therefore functionally operating as a Faster R-CNN model. This restricted the model to producing only bounding box predictions surrounding the pith. Ground truth bounding boxes were $20\times20$ pixel boxes surrounding the pith pixel location. For performance, initial training runs showed that Mask R-CNN struggled with overlapping bounding boxes, reducing its IoU scores. The model was trained with a batch size of 10, a learning rate of 0.004, momentum of 0.85, weight decay of 0.001, epochs set to 15, and a learning rate scheduler with a step size of 7 and $\gamma = 0.6$. The optimizer used was Stochastic Gradient Descent (SGD) optimizer, and a cross-entropy loss function was used. To mitigate the issue of overlapping detections, Non-Maximum Suppression (NMS) was applied using an IoU threshold of 0.5 during post-processing, improving the IoU score significantly.

\section{Results and Discussion}

This section presents a comprehensive analysis of the performance of the deep learning models evaluated for pith detection in tree cross-sections. The analysis is structured into three parts: quantitative performance comparison based on accuracy, precision, recall, F1-score, and IoU; training and validation loss analysis to understand convergence behavior and generalization; an exploratory analysis section to further analyze the model results; and an error analysis highlighting key challenges and model-specific limitations.

\subsection{Performance Comparison}

The models were assessed based on multiple performance metrics, including accuracy, precision, recall, F1-score, and Intersection over Union (IoU). These metrics provided a detailed assessment of how well each model performed in detecting and segmenting the pith region. The results are summarized in Table \ref{tab:performance}.

\begin{table}[htbp]
\centering
\resizebox{\linewidth}{!}{%
\begin{tabular}{|l|c|c|c|c|c|}
\hline
\textbf{Model} & \textbf{Accuracy} & \textbf{Precision} & \textbf{Recall} & \textbf{F1-Score} & \textbf{IoU}  \\
\hline
\textbf{Swin Transformer} & \textbf{0.94} & \textbf{0.93} & 0.92 & \textbf{0.93} & \textbf{0.85} \\
\textbf{YOLOv9}           & 0.92 & 0.91 & 0.89 & 0.90 & 0.82 \\
\textbf{DeepLabV3}       & 0.90 & 0.90 & 0.88 & 0.89 & 0.80 \\
\textbf{U-Net}           & 0.91 & 0.88 & 0.86 & 0.87 & 0.82 \\
\textbf{Mask R-CNN}      & N/A & 0.38 & \textbf{0.95} & 0.54 & 0.80 \\
\hline
\end{tabular}%
}
\caption{Comparative performance of five deep learning models on the primary image dataset,}
\label{tab:performance}
\end{table}

\textbf{Swin Transformer} achieved the highest accuracy of 0.94 and an IoU of 0.85, making it the most precise model for pith localization. The hierarchical feature learning and self-attention mechanisms allowed it to capture both local and global structural details of tree cross-sections effectively. However, despite its superior accuracy, it required significantly higher computational resources compared to other models, making real-time deployment challenging. 

\textbf{YOLOv9}, known for its real-time object detection capabilities, performed well with an accuracy of 0.92 and an IoU of 0.82. However, it struggled with fine boundary delineation, particularly when detecting pith regions with irregular edges. This limitation likely stems from its anchor-box-based approach, which is optimized for broader object localization rather than precise segmentation.

\textbf{DeepLabV3} demonstrated a balanced performance, achieving an accuracy of 0.90 and an IoU of 0.80. It effectively captured multi-scale features using its Atrous Spatial Pyramid Pooling (ASPP) mechanism, allowing it to adapt to variations in tree ring density. However, its reliance on dilated convolutions led to slight boundary imprecision in certain cases.

\textbf{U-Net} exhibited a strong balance between efficiency and segmentation performance, with an accuracy of 0.91 and an IoU of 0.82. Its encoder-decoder architecture enabled precise pith localization in structured patterns. However, it occasionally misclassified densely ringed areas as pith locations due to its tendency to over-rely on local features.

\textbf{Mask R-CNN}, due to only producing bounding boxes as a result of its mask being disabled, did not have an accuracy score. Regardless, its Recall score of 0.95 outperformed all other models while its IoU score of 0.80 was moderate. Additionally, the model initially exhibited the lowest IoU of 0.45 prior to Non-Maximum Suppression (NMS) with a threshold of 0.5 having been applied, leading to more precise and reliable object detection. A key issue was the presence of overlapping bounding boxes, which led to redundant predictions. This was reflected by the model's low Precision and F1-Score, both of which were the lowest scores of the four models tested, indicating that the model struggled with false positives, incorrectly detecting non-pith locations on tree cross-sections. 

\subsection{Exploratory Analysis}

Here, in order to further evaluate the the performance of the Swin Transformer model\textemdash{}the highest performing model of the five models\textemdash{}the model was used to test on the unseen oak tree specie dataset (consisting of 11 annotated images) from Oregon State University's Tree Ring Lab. This was done in order to assess the model's generalizability to unseen tree species.  

Additionally, its performance was compared to the Mask R-CNN which was chosen due to its low performance on the prior dataset consisting of 582 images to highlight the difference in performance between it and the Swin Transformer. 

During this model comparison, a different instance of the \textit{same} pretrained TorchVision Mask R-CNN model was trained \textit{solely} on a separate dataset of 64 tree cross-sections and finally tested on the 11 oak dataset images. (The 64 image dataset consists of tree cross-sections of \textit{Pinus taeda} trees from northern Uruguay \cite{marichal2024urudendropublicdatasetcrosssection}.) This was done in order to explore the potential generalizability performance improvements that would result from training with a smaller dataset. (Data augmentation was performed in a similar manner to how data augmentation was performed for the Mask R-CNN model trained on the 582 image dataset.) The Mask R-CNN models applied NMS post-processing to their results.  

\begin{figure}[htbp]
    \centering
    \raisebox{-.5\height}{\includegraphics[width=0.3\textwidth]{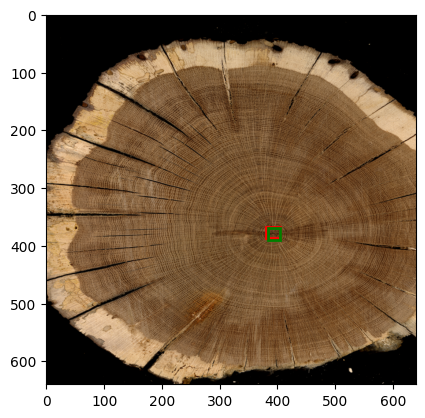}}%
    \hfill
    \raisebox{-.5\height}{\includegraphics[width=0.3\textwidth]{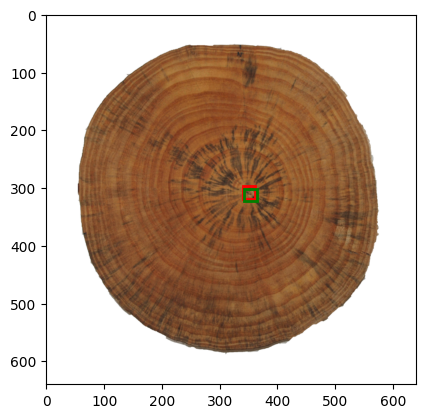}}%
    \caption{Oak dataset example tree cross-section for pith detection (top) and an example tree cross-section pith detection from the 582 images (bottom). The red box denotes the predicted bounding box while the green box denotes the ground truth bounding box.}
    \label{fig:side_by_side}
\end{figure}

As shown in Table \ref{tab:next1_performance}, the Mask R-CNN trained on the 64 image dataset outperformed the Mask R-CNN trained on 582 images and the Swin Transformer (without NMS applied) on the oak dataset. In contrast, in Table \ref{tab:next2_performance}, once Swin Transformer applied NMS during post-processing, it outperformed the other two models.

\begin{table}[H]
\centering
\resizebox{\linewidth}{!}{%
\begin{tabular}{|l|c|c|c|c|c|}
\hline
\textbf{Model} & \textbf{Accuracy} & \textbf{Precision} & \textbf{Recall} & \textbf{F1-Score} & \textbf{IoU}  \\
\hline
\textbf{Swin Transformer (No NMS)} & 0.68 & 0.62 & 0.59 & 0.60 & 0.50 \\
\textbf{Mask R-CNN (64 images)}    & N/A & \textbf{0.70} & \textbf{0.64} & \textbf{0.67} & \textbf{0.60} \\
\textbf{Mask R-CNN (582 images)}   & N/A & 0.39 & 0.64 & 0.48 & 0.56 \\
\hline
\end{tabular}%
}
\caption{ Performance of selected models (Swin Transformer and Mask R-CNN) on the oak dataset without applying Non-Maximum Suppression (NMS) to the Swin Transformer, highlighting its baseline generalization to unseen hardwood data.}
\label{tab:next1_performance}
\end{table}

\begin{table}[H]
\centering
\resizebox{\linewidth}{!}{%
\begin{tabular}{|l|c|c|c|c|c|}
\hline
\textbf{Model} & \textbf{Accuracy} & \textbf{Precision} & \textbf{Recall} & \textbf{F1-Score} & \textbf{IoU}  \\
\hline
\textbf{Swin Transformer (Using NMS)} & \textbf{0.87} & \textbf{0.83} & \textbf{0.80} & \textbf{0.81} & \textbf{0.75} \\
\textbf{Mask R-CNN (64 images)}    & N/A & 0.70 & 0.64 & 0.67 & 0.60 \\
\textbf{Mask R-CNN (582 images)}   & N/A & 0.39 & 0.64 & 0.48 & 0.56 \\
\hline
\end{tabular}%
}
\caption{Performance of selected models on the oak dataset after applying NMS to the Swin Transformer, showing improved generalization metrics, especially for Swin Transformer.}
\label{tab:next2_performance}
\end{table}

\subsection{Training and Validation Loss Analysis}

To complement the quantitative performance metrics, the loss curves of each model was analyzed to assess convergence behavior, generalization capability, and overfitting risks.

\begin{figure}[htbp]
    \centering
    \includegraphics[width=0.9\linewidth]{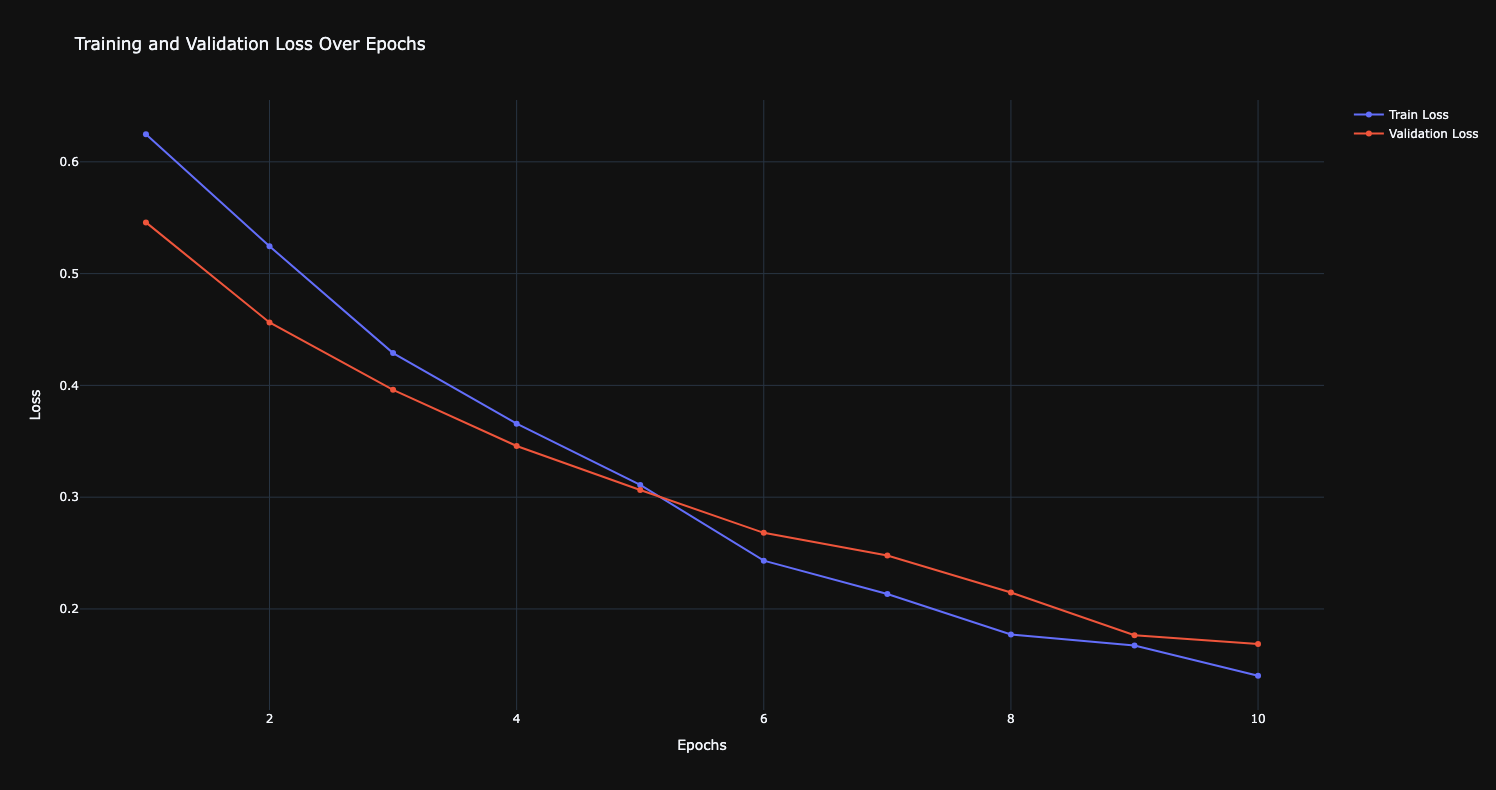}
    \caption{Training and Validation Loss Curve for Swin Transformer.}
    \label{fig:swin_loss}
\end{figure}

The \textbf{Swin Transformer} loss curve in Figure \ref{fig:swin_loss} indicates a slower convergence compared to other models. This is expected due to the computational complexity of transformer-based architectures, which require a large amount of training data to generalize effectively. Although it ultimately achieved the best accuracy, its training time was significantly longer.

\begin{figure}[htbp]
    \centering
    \includegraphics[width=0.9\linewidth]{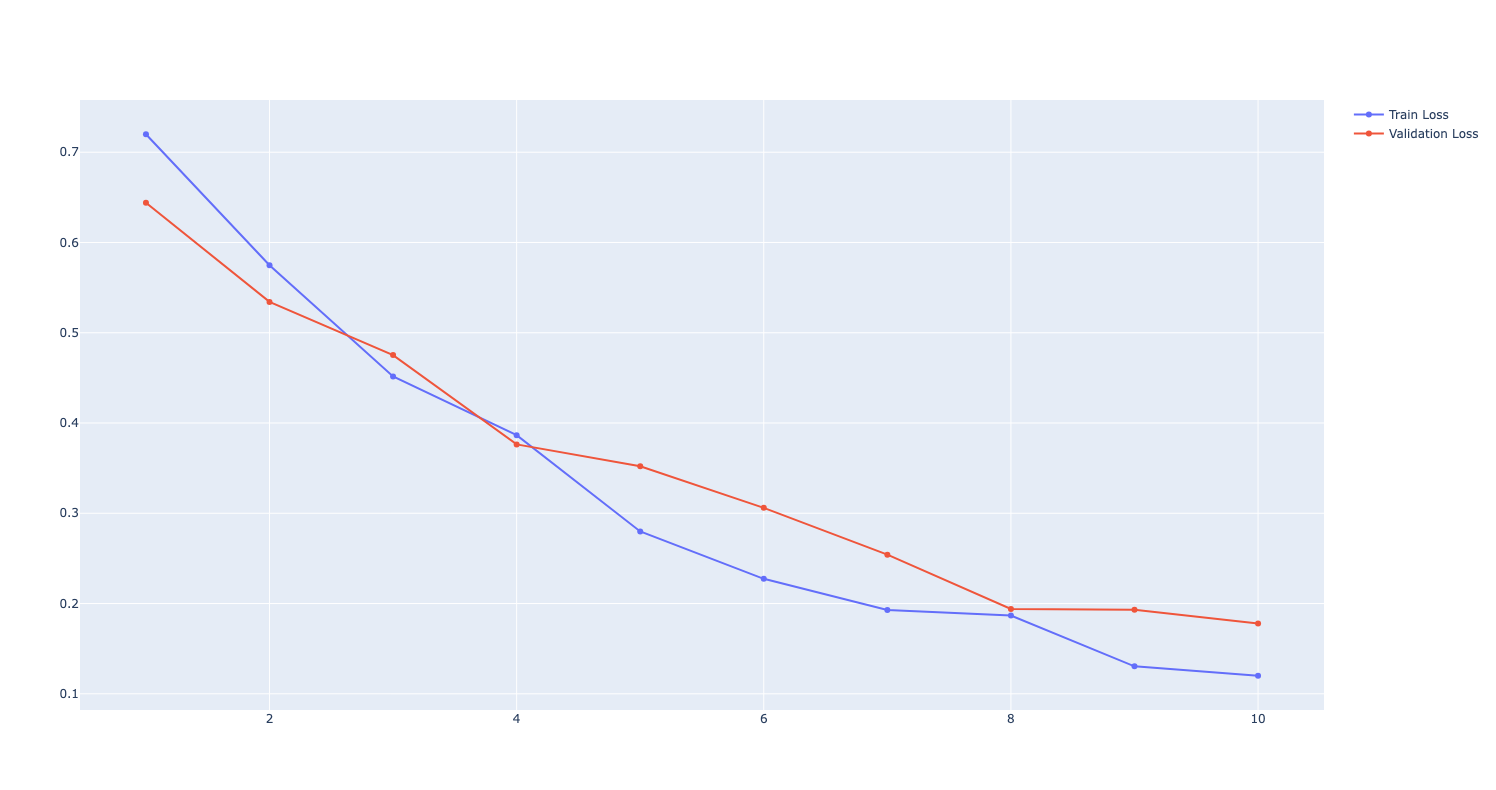}
    \caption{Training and Validation Loss Curve for YOLOv9.}
    \label{fig:yolo_loss}
\end{figure}

Figure \ref{fig:yolo_loss} shows that \textbf{YOLOv9}'s training loss decreased rapidly, but its validation loss plateaued early. This suggests that while the model efficiently learned general object detection, it did not refine boundary delineation effectively. This aligns with its performance in Table \ref{tab:performance}, where it struggled with complex pith structures.

\begin{figure}[H]
    \centering
    \includegraphics[width=0.9\linewidth]{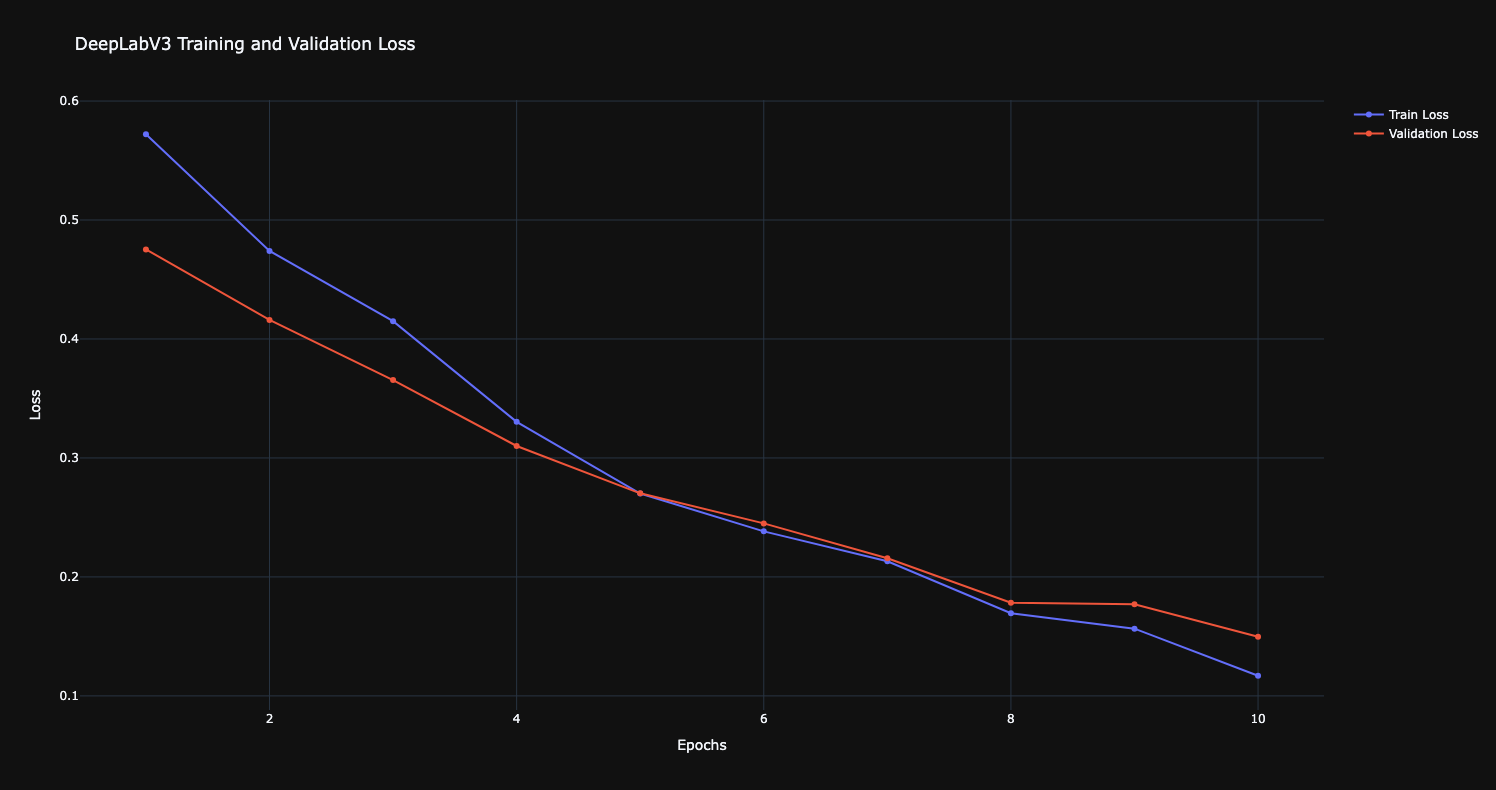}
    \caption{Training and Validation Loss Curve for DeepLabV3.}
    \label{fig:deeplab_loss}
\end{figure}

\textbf{DeepLabV3} exhibited a stable loss curve with minimal oscillations, as seen in Figure \ref{fig:deeplab_loss}. This indicates smooth learning with minimal overfitting. However, its reliance on dilated convolutions led to minor inaccuracies in detecting fine segmentation boundaries.

\begin{figure}[H]
    \centering
    \includegraphics[width=0.9\linewidth]{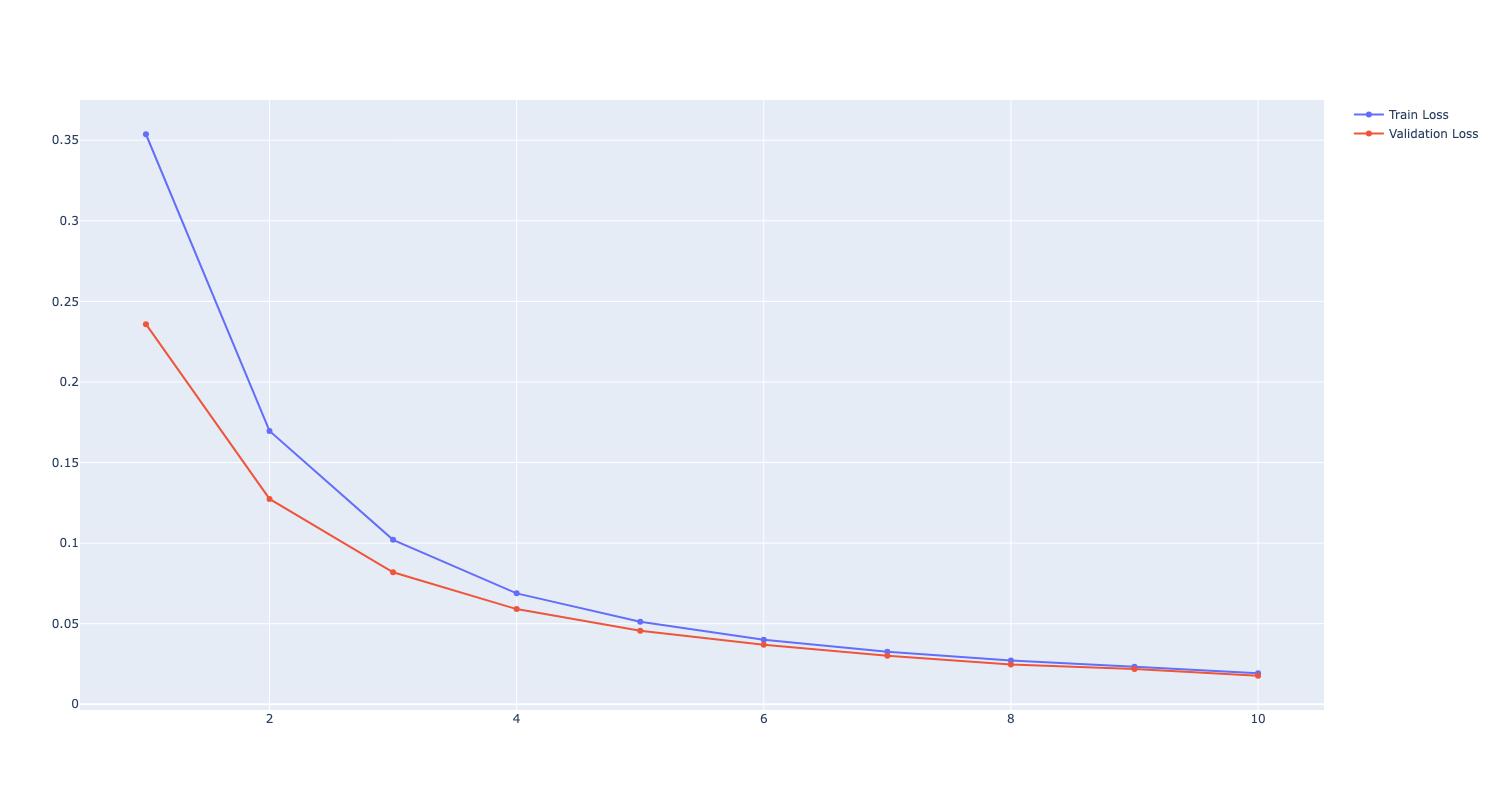}
    \caption{Training and Validation Loss Curve for U-Net.}
    \label{fig:unet_loss}
\end{figure}

\textbf{U-Net}, shown in Figure \ref{fig:unet_loss}, maintained good generalization across the dataset. However, occasional fluctuations in loss indicate sensitivity to texture variations in tree cross-sections, which resulted in false positives in complex structures.

\begin{figure}[htbp]
    \centering
    \includegraphics[width=0.9\linewidth]{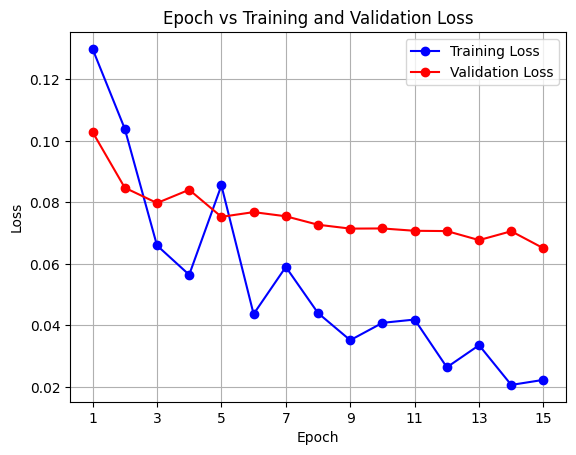}
    \caption{Training and Validation Loss Curve for Mask R-CNN.}
    \label{fig:maskrcnn_loss}
\end{figure}

Figure \ref{fig:maskrcnn_loss} highlights the high fluctuations in \textbf{Mask R-CNN’s} training loss, suggesting instability due to redundant bounding boxes. The implementation of NMS helped stabilize predictions, although a noticeable gap between the training loss and validation loss was evident during training.

\subsection{Error Analysis}
Several challenges were encountered during model training and evaluation. One notable issue was the presence of overlapping detections in Mask R-CNN, leading to performance degradation. To resolve this, Non-Maximum Suppression (NMS) was implemented, filtering out redundant bounding boxes and improving IoU from 0.45 to 0.80.

Another key observation was that YOLOv9, while excelling at object detection, struggled with precise boundary delineation. This led to minor errors in cases where the pith had irregular edges or diffuse boundaries. Further refinements in the anchor box configurations could potentially improve this issue.

U-Net exhibited misclassifications in tree cross-sections with complex textures. The model sometimes mistook densely ringed areas as pith locations, leading to false positives. This issue was partially mitigated through additional data augmentation techniques to expose the model to a wider range of structural variations.

\section{Conclusion}

This study presented a comprehensive evaluation of deep learning models for automated pith detection in tree cross-sections. Among the five models explored—Swin Transformer, YOLOv9, DeepLabV3, U-Net, and Mask R-CNN—the Swin Transformer achieved the highest accuracy and segmentation precision, demonstrating its suitability for fine-grained localization tasks. YOLOv9 offered superior inference speed but exhibited reduced accuracy in complex boundary cases. DeepLabV3 and U-Net balanced accuracy with computational efficiency, while Mask R-CNN benefited significantly from Non-Maximum Suppression, improving its IoU from 0.45 to 0.80.

Beyond quantitative metrics, loss curve analysis revealed each model’s convergence behavior and generalization capability. Swin Transformer and DeepLabV3 showed stable learning, while YOLOv9 and Mask R-CNN displayed signs of early plateauing and fluctuations, respectively. Overall, our findings underscore that model selection should consider both task-specific accuracy and deployment constraints such as real-time performance and hardware resources.

\section{Future Work}

Future research will focus on reducing the computational cost of transformer-based models through pruning, quantization, or knowledge distillation to enable deployment in real-time environments. Hybrid models that integrate detection and segmentation components—such as combining YOLOv9 with U-Net—may improve boundary precision without sacrificing speed. Additionally, expanding the dataset to include more species and imaging conditions will enhance model robustness. For instance, the finding that the Mask R-CNN generalized better to the oak dataset when trained using the 64 image dataset highlights the importance of dataset quality. Lastly, integrating attention mechanisms into CNN architectures and testing models in real-world forestry settings will be key to bridging the gap between lab-scale accuracy and practical utility.

{\small
\bibliographystyle{ieee}
\bibliography{egbib}

@article{schraml2013pith,
  author = {Schraml, R. and Uhl, A.},
  title = {Pith estimation on rough log end images using local fourier spectrum analysis},
  journal = {Computers, Graphics, and Imaging},
  volume = 35,
  number = 4,
  pages = {245--258},
  year = 2013
}

@article{norell2009automatic,
  author = {Norell, K.},
  title = {An automatic method for counting annual rings in noisy sawmill images},
  journal = {International Conference on Image Analysis and Processing (ICIAP)},
  volume = 29,
  number = 3,
  pages = {115--130},
  year = 2009
}

@article{kurdthongmee2018fast,
  author = {Kurdthongmee, W.},
  title = {A fast algorithm to approximate the pith location of rubberwood timber from a normal camera image},
  journal = {Journal of Computer Science and Software Engineering},
  volume = 45,
  number = 2,
  pages = {78--91},
  year = 2018
}

@article{decelle2022ant,
  author = {Decelle, R. and colleagues},
  title = {Ant Colony Optimization for Estimating Pith Position on Images of Tree Log Ends},
  journal = {Image Processing On Line},
  volume = 11,
  pages = {512--526},
  year = 2022
}

@article{kurdthongmee2020comparative,
  author = {Kurdthongmee, W.},
  title = {A comparative study of the effectiveness of using popular DNN object detection algorithms for pith detection in cross-sectional images of parawood},
  journal = {Heliyon},
  volume = 6,
  number = 10,
  pages = {e04879},
  year = 2020
}

@article{ronneberger2015u,
  author = {Ronneberger, O. and Fischer, P. and Brox, T.},
  title = {U-Net: Convolutional networks for biomedical image segmentation},
  journal = {Medical Image Computing and Computer-Assisted Intervention (MICCAI)},
  volume = 9351,
  pages = {234--241},
  year = 2015
}

@article{he2017mask,
  author = {He, K. and Gkioxari, G. and Dollár, P. and Girshick, R.},
  title = {Mask R-CNN},
  journal = {International Conference on Computer Vision (ICCV)},
  volume = 30,
  number = 2,
  pages = {2980--2988},
  year = 2017
}

@article{liu2021swin,
  author = {Liu, Z. and Lin, Y. and Cao, Y. and Hu, H. and Wei, Y. and Zhang, Z. and Lin, S. and Guo, B.},
  title = {Swin Transformer: Hierarchical vision transformer using shifted windows},
  journal = {International Conference on Computer Vision (ICCV)},
  volume = 34,
  number = 1,
  pages = {10012--10022},
  year = 2021
}

@article{chen2017deeplab,
  author = {Chen, L. C. and Papandreou, G. and Kokkinos, I. and Murphy, K. and Yuille, A. L.},
  title = {Rethinking atrous convolution for semantic image segmentation},
  journal = {arXiv preprint arXiv:1706.05587},
  year = 2017
}

@article{DBLP:journals/corr/abs-2404-01952,
  author       = {Henry Marichal and
                  Diego Passarella and
                  Gregory Randall},
  title        = {Automatic Wood Pith Detector: Local Orientation Estimation and Robust
                  Accumulation},
  journal      = {CoRR},
  volume       = {abs/2404.01952},
  year         = {2024},
  url          = {https://doi.org/10.48550/arXiv.2404.01952},
  doi          = {10.48550/ARXIV.2404.01952},
  eprinttype    = {arXiv},
  eprint       = {2404.01952},
  bibsource    = {dblp computer science bibliography, https://dblp.org}
}

@misc{marichal2024urudendropublicdatasetcrosssection,
      title={UruDendro, a public dataset of cross-section images of Pinus taeda}, 
      author={Henry Marichal and Diego Passarella and Christine Lucas and Ludmila Profumo and Verónica Casaravilla and María Noel Rocha Galli and Serrana Ambite and Gregory Randall},
      year={2024},
      eprint={2404.10856},
      archivePrefix={arXiv},
      primaryClass={cs.CV},
      url={https://arxiv.org/abs/2404.10856}, 
}
}

\end{document}